\documentclass{article}
\usepackage{spconf,amsmath,amssymb,graphicx,booktabs}
\usepackage[numbers,sort&compress]{natbib}
\usepackage{hyperref}
\usepackage{eso-pic}
\usepackage{textcomp}
\usepackage[numbers,sort&compress]{natbib}

\title{SIE3D: Single-Image Expressive 3D Avatar Generation via Semantic Embedding and Perceptual Expression Loss}
\name{Zhiqi Huang\sthanks{Corresponding Author}, Dulongkai Cui, Jinglu Hu}
\address{Graduate School of Information, Production and Systems\\
Waseda University\\
Kitakyushu, Japan\\
\{zhiqi.huang, dulongkai.cui\}@akane.waseda.jp jinglu@waseda.jp}
\begin{document}
\ninept
\maketitle

\AddToShipoutPictureFG*{%
  \AtPageLowerLeft{%
    \hspace{0.70in}%
    \raisebox{0.32in}{%
      \parbox[b]{7.10in}{\tiny
      \textcopyright~2026 IEEE. Personal use of this material is permitted.
      Permission from IEEE must be obtained for all other uses, in any current or future media,
      including reprinting/republishing this material for advertising or promotional purposes,
      creating new collective works, for resale or redistribution to servers or lists,
      or reuse of any copyrighted component of this work in other works.\\
      Published in the 2026 IEEE International Conference on Acoustics, Speech and Signal Processing (ICASSP),
      pp.~9897--9901. DOI:
      \href{https://doi.org/10.1109/ICASSP55912.2026.11462135}{10.1109/ICASSP55912.2026.11462135}.%
      }%
    }%
  }%
}

\begin{abstract}
Generating high-fidelity 3D head avatars from a single image is challenging,
as current methods lack fine-grained, intuitive control over expressions 
via text. This paper proposes SIE3D, a framework that generates expressive 
3D avatars from a single image and descriptive text. SIE3D fuses identity 
features from the image with semantic embedding from text through a novel 
conditioning scheme, enabling detailed control. To ensure generated 
expressions accurately match the text, it introduces an innovative 
perceptual expression loss function. This loss uses a pre-trained expression 
classifier to regularize the generation process, guaranteeing expression 
accuracy. Extensive experiments show SIE3D significantly improves 
controllability and realism, outperforming competitive methods in 
identity preservation and expression fidelity on a single consumer-grade GPU. 
Project page: https://huang-zhiqi.github.io/SIE3D/
\end{abstract}
\section{Introduction}
\label{sec:intro}

Creating realistic and animatable 3D digital human avatars is crucial for virtual 
reality, filmmaking, and other applications \cite{chu2024gpavatar, chu2024generalizable, wang2024survey}. A significant challenge is "one-shot" 
generation, which aims to create a complete 3D head model from a single photograph \cite{gerogiannis2025arc2avatar, han2023headsculpt}. 
This problem is inherently difficult, as it requires inferring full 3D geometry and 
appearance from limited 2D information. The core issue in current research is 
balancing two objectives: identity fidelity, ensuring the avatar accurately 
resembles the person, and editability, allowing users to intuitively modify 
attributes like facial expressions. Existing approaches often excel at one but 
not both \cite{wang2024survey, canfes2023text, wang20253d, li2024advances}.

Avatar generation technology has progressed from traditional 3D deformable models 
(3DMMs) \cite{blanz2023morphable, li2017learning} to modern neural network-based implicit representations like Neural 
Radiance Fields (NeRFs) \cite{mildenhall2021nerf} and explicit ones like 3D Gaussian Splatting (3DGS) \cite{kerbl20233d}. 
Arc2Avatar \cite{gerogiannis2025arc2avatar}, a leading one-shot method, has achieved breakthroughs in identity 
fidelity by using the powerful 2D face model Arc2Face \cite{papantoniou2024arc2face} as a prior and combining 
it with the high-quality rendering of 3DGS. Despite its success, Arc2Avatar has 
key limitations. First, its expression control is non-semantic; it achieves 
expression variations by driving the blendshapes of an underlying FLAME \cite{li2017learning} parametric 
model, a type of 3DMM. This requires users to manipulate abstract numerical 
parameters rather than using natural language commands. Second, it suffers from 
stability issues and can occasionally fail to produce a neutral pose, a known 
limitation from relying solely on identity priors.

Meanwhile, an independent research direction, with representative works including DreamFusion \cite{poole2022dreamfusion},
HeadSculpt \cite{han2023headsculpt} and  TADA \cite{liao2024tada}, focuses on generating 3D head portraits from text 
descriptions. While these methods offer a high degree of creativity and semantic 
control, they cannot faithfully reconstruct a specific person's identity from a 
photo. This leads to a dichotomy in the field: image-to-3D methods like Arc2Avatar 
preserve identity but lack semantic editing, while text-to-3D methods like 
HeadSculpt offer semantic control but cannot maintain a specific identity. 
Furthermore, the few existing methods \cite{canfes2023text, sun2023dreamcraft3d} that accept both image and text are 
rare and generally yield unsatisfactory results \cite{wang2024survey}. There 
is currently no unified solution that achieves both.

This paper aims to bridge this gap. Our proposed SIE3D framework is designed to 
combine the high-fidelity identity preservation of image reconstruction with the 
powerful semantic controllability of text-based models. The core idea is a novel 
multimodal conditioning mechanism that uses dual inputs to guide generation. 
Identity embedding from a single image ensure the avatar's structure is consistent 
with the subject, while semantic embedding from text prompts control its state, 
such as expression and other attributes. To ensure the generated expressions 
faithfully match the text, we also introduce a perceptual expression regularization loss 
function, which leverages a pre-trained facial expression classifier to guide 
the 3D model's optimization in a semantically rich feature space.

The main contributions can be summarized as follows:
\begin{itemize}
  \item We propose a novel multimodal framework, SIE3D, that generates 3D avatars 
  from a single image and descriptive text. For the first time, it enables 
  fine-grained, semantic control of expressions and attributes while 
  maintaining a high level of identity consistency.
  \item We introduce a disentangled conditioning mechanism that fuses independent 
  expression and edit embedding, enabling combined control of an avatar's facial 
  expressions and appearance attributes via natural language.In addition, we introduce an expression-aware loss function 
  that innovatively leverages a pretrained face analysis model as a semantic 
  regularizer, significantly improving the accuracy and realism of generated 
  expressions and effectively bridging the semantic gap in the SDS loss.
  \item Extensive experiments on a single consumer-grade GPU demonstrate that our 
  approach performs comparably with competitive methods in generating 
  expressive 3D avatars with high identity fidelity.
\end{itemize}

\begin{figure*}[t]
  \centering
  \centerline{\includegraphics[width=15cm]{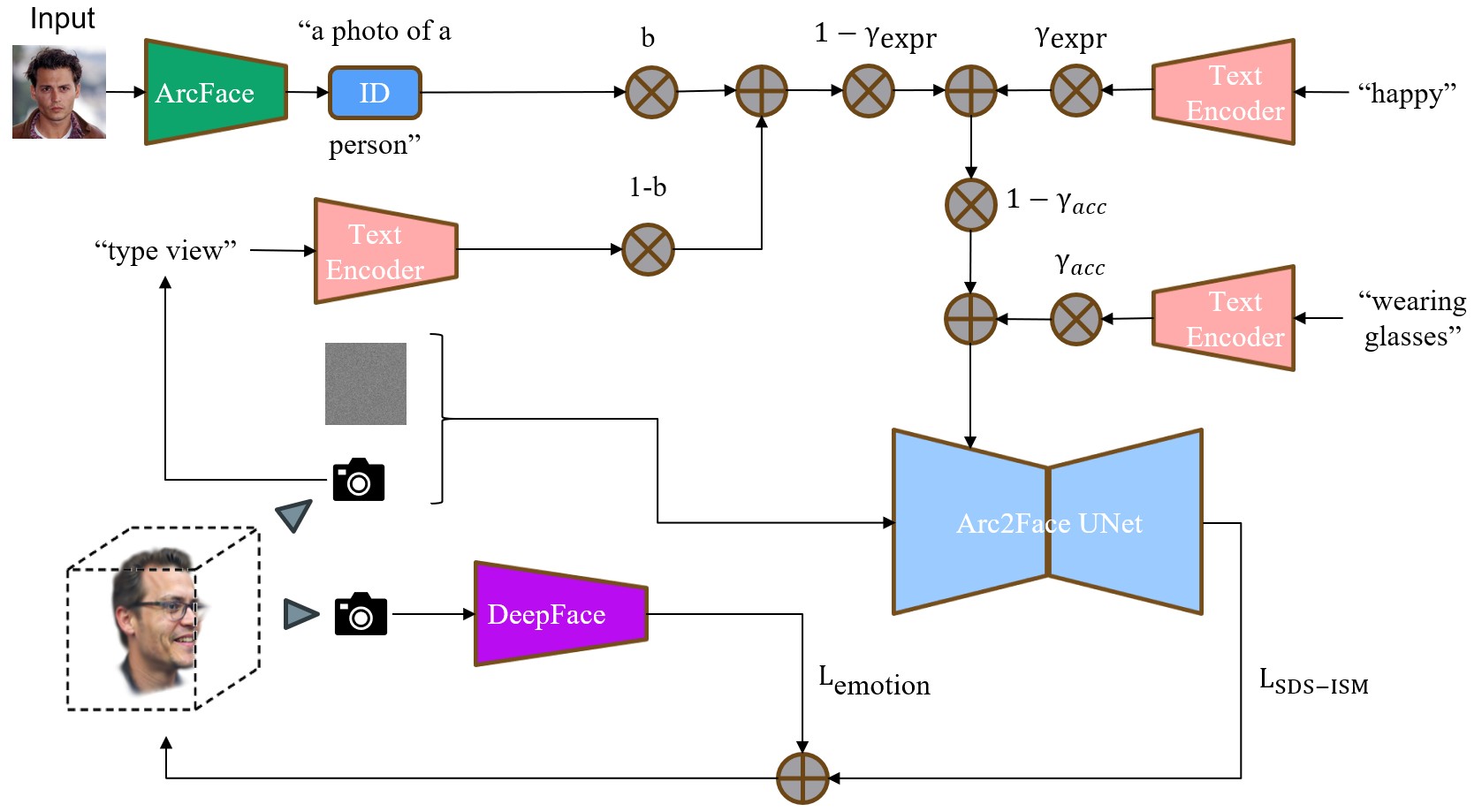}}
\caption{ \textbf{Overall architecture of the SIE3D framework.} The model takes a single 
image and text prompts (e.g., "happy," "wearing glasses") as multimodal inputs. It 
fuses identity embedding extracted by ArcFace \cite{deng2019arcface} with semantic embedding from a Stable Diffusion \cite{rombach2022high} CLIP text 
encoder \cite{radford2021learning} via a hierarchical conditioning mechanism to jointly guide the Arc2Face UNet \cite{gerogiannis2025arc2avatar}. 
To ensure expression accuracy, a perceptual expression regularization loop is introduced: the 
rendered output from the model is fed into a pre-trained DeepFace \cite{taigman2014deepface} model to compute 
an perceptual expression loss ($\mathcal{L}_{emotion}$). This loss, combined with the primary 
SDS-ISM \cite{liang2024luciddreamer} loss, jointly optimizes the final 3D representation.}
\label{fig:res1}
\end{figure*}

\section{Method}
\label{sec:method}

\subsection{Method Overview}
\label{ssec:method overview}

As shown in Figure 1, we propose a novel framework using multi-layer semantic embedding and perceptual expression loss.
Our method aims to generate highly expressive 3D head avatars from 
a single image. We build upon the robust framework of Arc2Avatar, which leverages a 
fine-tuned 2D face foundation model to guide the optimization of a 3D Gaussian 
Splatting (3DGS) representation via Score Distillation Sampling (SDS) \cite{ruiz2023dreambooth}. Our primary 
contributions enhance this framework by introducing two key innovations. First, we 
extend the conditioning mechanism beyond identity and viewpoint to include explicit 
controls for facial expressions and accessories. This is achieved by creating and 
composing dedicated text embedding, allowing for fine-grained semantic manipulation 
of the generated avatar. Second, to ensure the fidelity of the generated expressions, 
we introduce a novel perceptual expression regularization term. This term utilizes the DeepFace \cite{taigman2014deepface}
model to analyze the expression of the rendered avatar and computes a cross-entropy 
loss against the target expression, thereby enforcing greater consistency and accuracy 
during optimization. Our final framework enables the creation of avatars that not only 
preserve the subject's identity but also offer a high degree of artistic control over 
their expressions and appearance.

\subsection{Preliminary}
\label{ssec:preliminary}

Our work is fundamentally based on the Arc2Avatar  methodology, which generates a 3D 
head avatar represented by 3D Gaussian Splats, $\mathcal{G}$, parameterized by 
$\theta$. The optimization process is guided by a 2D diffusion model using an 
advanced Score Distillation Sampling (SDS) technique.

{\bf ID-Conditioned Guidance via ISM.} The core of the generation process is guided by 
a powerful, identity-aware 2D diffusion model, an augmented version of Arc2Face. 
Instead of the standard SDS, Arc2Avatar employs Interval Score Matching (ISM) \cite{liang2024luciddreamer}  for 
its superior stability and ability to generate high-fidelity results. The ISM loss 
gradient, which optimizes the 3D representation $\theta$, is defined as:
\begin{equation}\nabla_\theta\mathcal{L}_{ISM}(\theta)=\mathbb{E}_{t,c,\epsilon}
  [w(t)||\epsilon_\phi(x_t,t,c)-\epsilon_\phi(x_s,s,\emptyset)||^2\nabla_\theta 
  f(\theta,c)]\end{equation}
where $f(\theta,c)$ is the differentiable rendering function that produces a 2D image 
$x_0$ from the 3D Gaussians $\mathcal{G}$ given camera parameters $c$.

$\epsilon_\phi$ is the denoising U-Net conditioned on embedding $c$, and $x_t$ and
$x_s$ are differently noised versions of the rendered image.

{\bf View-Enriched Conditioning.} The conditioning embedding $c$ is a crucial 
component that fuses identity and viewpoint information. The identity is captured by 
an ArcFace \cite{deng2019arcface} embedding $v$ from the input image. This embedding is integrated into a 
default text prompt to create an identity-conditioned embedding, $c_{default}$. To guide
the generation from various angles, this is blended with view-specific text embedding
($c_{view}$) obtained from the original Stable Diffusion \cite{rombach2022high} text encoder. The final 
view-enriched embedding $c_d$ for a given direction $d$ is computed as a linear 
interpolation:
\begin{equation}c_d=b\cdot c_{default}+(1-b)\cdot c_{view}\end{equation}
where $b\in [0,1]$ is a blending factor that balances identity preservation and view 
guidance.

{\bf Mesh-Based Regularization.} To maintain a coherent facial structure and enable 
blendshape-based animation, the optimization is regularized to adhere to an 
underlying FLAME \cite{li2017learning} mesh template. This is achieved through two geometric regularizers
: a positional L2 regularizer ($\mathcal{L}_{pos}$) that minimizes the distance 
between splat positions and their corresponding template vertices, and a Laplacian 
regularizer ($\mathcal{L}_{lap}$ that preserves the local geometric structure. The 
combined loss for the Arc2Avatar framework is thus:
\begin{equation}\mathcal{L}_{Arc2Avatar}=\mathcal{L}_{ISM}+\lambda_{pos}\mathcal{L}_
  {pos}+\lambda_{lap}\mathcal{L}_{lap}\end{equation}
where $\lambda_{pos}$ and $\lambda_{lap}$ are the weights for the respective 
regularization terms.

\subsection{Expressive Generation with Semantic Embedding}
\label{ssec:expressive generation with semantic embedding}

A key limitation of the original framework is its lack of explicit control over 
semantic attributes like facial expressions or accessories. We address this by 
introducing a hierarchical embedding composition strategy that extends the 
conditioning mechanism.

{\bf Multi-Attribute Embedding Generation.} We first compute a dictionary of 
embedding for various expressions (e.g., 'happy', 'sad', 'neutral') and 
accessories (e.g., 'wearing glasses', 'with beard'). Similar to the view embedding, 
these are generated by blending the base identity-conditioned embedding $c_{default}$
with attribute-specific text embedding from the SD text encoder. For a given 
expression $i$ or accessory $j$, the embedding are created as:
\begin{equation}c_{expr_i}=(1-\gamma_{expr})\cdot c_{default}+\gamma_{expr}
  \cdot E_{SD}(\text{"expression }i\mathrm{"})\end{equation}
\begin{equation}c_{acc_j}=(1-\gamma_{acc})\cdot c_{default}+\gamma_{acc}
  \cdot E_{SD}(\text{"accessory }j")\end{equation}
where $E_{SD}$ is the text encoder, and $gamma_{expr}$ and $gamma_{acc}$ are 
factors controlling the influence of the attribute text.

{\bf Hierarchical and Intensity-Aware Conditioning.} During the SDS optimization, 
we construct the final conditioning embedding $c_{final}$ in a hierarchical 
fashion. First, the view-enriched embedding $c_d$  is computed as in the base model. 
This embedding then serves as the new base for incorporating expression, which is 
subsequently used as a base for adding accessories. The control is made continuous 
through an intensity parameter $\eta \in [0,1]$. For a target expression 
$c_{target\_expr}$ with intensity $\eta_{expr}$, the expression-conditioned embedding
$c_{expr\_final}$ is interpolated from a neutral state:
\begin{equation}c_{expr\_final}=(1-\eta_{expr})\cdot c_{neutral}+\eta_{expr}\cdot 
  c_{target\_expr}\end{equation}
The final composite embedding $c_{final}$ is formed by sequentially applying these 
interpolations, allowing for smooth and disentangled control over viewpoint, 
expression, and accessories.

\subsection{Expression Regularization with Perceptual Loss}
\label{ssec:expression regularization with perceptual loss}

While the extended embedding guide the model towards a target expression, there is 
no mechanism to enforce its accurate depiction. To this end, we introduce a perceptual expression 
loss that explicitly regularizes the facial expression during optimization.

{\bf DeepFace-based Expression Analysis.} At each training iteration where the 
regularization is active, we render a frontal view $x_{frontal}$ of the 3D avatar. 
We then utilize the DeepFace model, a robust pre-trained facial analysis model, 
to obtain a predicted probability distribution $P_{pred}$ over a set of $k$ 
discrete emotion categories (e.g., 'angry', 'happy', 'neutral').
\begin{equation}P_{pred}=\text{DeepFace.analyze}(x_{frontal})=
  \{p_1,p_2,\ldots,p_k\}\end{equation}
where $p_i$ is the predicted probability for the $i$-th emotion and 
$\sum_{i=1}^k p_i=1$.

{\bf Cross-Entropy Regularization.} We define a target distribution, $P_{target}$, 
as a one-hot vector where the entry corresponding to the desired target emotion 
is 1 and all others are 0. We then compute the cross-entropy loss between the 
predicted and target distributions. This loss, denoted as $\mathcal{L}_{emotion}$,
penalizes deviations from the target expression:
\begin{equation}\mathcal{L}_{emotion}=-\sum_{i=1}^{k}P_{target,i}
  \log(P_{pred,i})\end{equation}
For instance, during the generation of a neutral avatar, this loss effectively 
minimizes any unintentional expressions, ensuring better correspondence with the 
neutral mesh template.

Final Loss Function. This perceptual expression loss is integrated into the main optimization 
objective. The final loss function for our proposed method is a weighted sum of the 
ISM loss, the geometric regularizers, and our new expression regularization term:
\begin{equation}{\mathcal{L}}_{final}={\mathcal{L}}_{ISM}+\lambda_{pos}
  {\mathcal{L}}_{pos}+\lambda_{lap}{\mathcal{L}}_{lap}+\lambda_{emotion}
  {\mathcal{L}}_{emotion}\end{equation}
This combined objective function guides the generation of an identity-consistent 
3D avatar that is also faithful to the desired viewpoint, expression, and accessory 
attributes.

\begin{figure*}[t]
  \centering
  \centerline{\includegraphics[width=16cm]{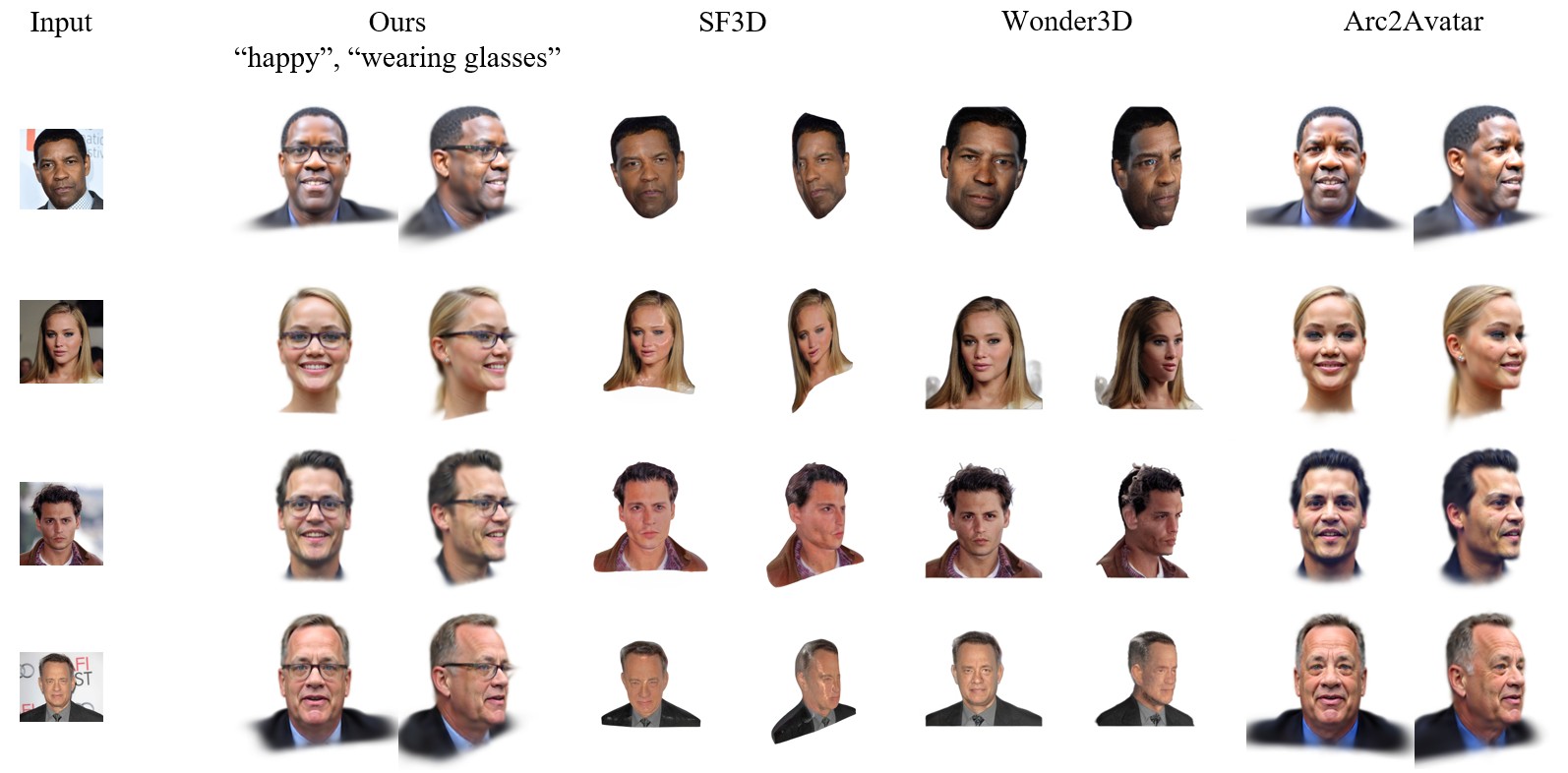}}
\caption{\textbf{Qualitative comparison with other competitive methods.} 
Qualitative comparison shows our method's superiority in semantic controllability 
and multi-view identity preservation over other competitive approaches.}
\label{fig:res2}
\end{figure*}

\section{Experiments}
\label{sec:experiments}

\subsection{Implementation Details}
\label{ssec:implementation details}
Our experiments were conducted using the Celebrity Face Image Dataset \cite{thakur2022_celebrity_face_kaggle}, with all 
images uniformly resized to a 512x512 resolution. All models were implemented in 
PyTorch and trained on a single NVIDIA RTX 4090 GPU. Each avatar was optimized for 
5000 iterations, a process that took approximately 70 minutes.

\subsection{Quantitative Results}
\label{ssec:quantitative results}
Due to the limited ability of text-to-3D methods to generate meaningful 3D identities and limited space, we exclude them from the comparison.
We quantitatively compare our method against three competitive image-to-3D approaches: 
Arc2Avatar \cite{gerogiannis2025arc2avatar}, Wonder3D \cite{long2024wonder3d}, and SF3D \cite{boss2025sf3d}. The evaluation was based on three metrics: 
Fréchet Inception Distance (FID) \cite{heusel2017gans} to measure realism, Identity Consistency (ID) 
calculated using the CosFace similarity metric \cite{wang2018cosface}, and our proposed Neutrality Preservation Score (NPS) to measure 
stability in maintaining a neutral expression via cross-entropy loss. To ensure a 
fair comparison, our method takes an image and a default text prompt ("no accessories, neutral 
expression") as input, while the other methods use an image. As shown in Table 1, 
our method achieves the best performance in realism (FID) and neutrality 
preservation (NPS). While the baseline Arc2Avatar achieves the highest ID score, 
indicating the best identity preservation, our method maintains a 
competitive ID score that is better than other compared methods.

\begin{table}[htbp]
  \caption{Quantitative Comparison Results}
  \centering
  \begin{tabular*}{\columnwidth}{c @{\extracolsep{\fill}} c c c} 
    \toprule
    Method & FID$\downarrow$ & ID$\uparrow$ & NPS$\downarrow$\\
    \midrule
    Wonder3D \cite{long2024wonder3d} & 274.10 & 0.2603 & 0.4621\\
    SF3D \cite{boss2025sf3d} & 290.89 & 0.3187 & 2.9998\\
    Arc2Avatar \cite{gerogiannis2025arc2avatar} & 237.81 & \textbf{0.4459} & 3.5884\\
    Ours & \textbf{227.11} & 0.3672 & \textbf{0.3667}\\
    \bottomrule
  \end{tabular*}
\end{table}

\subsection{Qualitative Results}
\label{ssec:qualitative results}

As shown in Figure 2, we conducted a qualitative comparison of the generated 
results. Our method introduces powerful semantic editing capabilities while maintaining identity consistency. In contrast, although 
Wonder3D and SF3D can generate high-quality frontal images, they fail to preserve 
quality in side-view perspectives.

\subsection{Ablation Study}
\label{ssec:ablation study}

We conducted an ablation study to validate the effectiveness of our two key proposed 
components: "semantic embedding" and "perceptual expression loss". As shown in Table 2, 
removing the "perceptual expression loss" severely degrades performance across all 
metrics, highlighting its critical role in the framework. Interestingly, removing 
"semantic embedding" slightly improves the ID score but at a significant cost to both 
realism (FID) and neutrality preservation (NPS). This demonstrates a trade-off and 
validates the importance of semantic embedding for achieving high overall quality 
and expression control.

\begin{table}[htbp]
  \caption{Ablation Study Results}
  \centering
  \begin{tabular*}{\columnwidth}{c @{\extracolsep{\fill}} c c c} 
    \toprule
    Method & FID$\downarrow$ & ID$\uparrow$ & NPS$\downarrow$\\
    \midrule
    full & \textbf{227.11} & 0.3672 & \textbf{0.3667}\\
    w/o semantic embedding & 238.43 & \textbf{0.3977} & 3.1178\\
    w/o perceptual expression loss & 321.68 & 0.2539 & 3.8043\\
    \bottomrule
  \end{tabular*}
\end{table}

\subsection{Applications}
\label{ssec:applications}

\begin{figure}[t]
  \centering
  \centerline{\includegraphics[width=8.5cm]{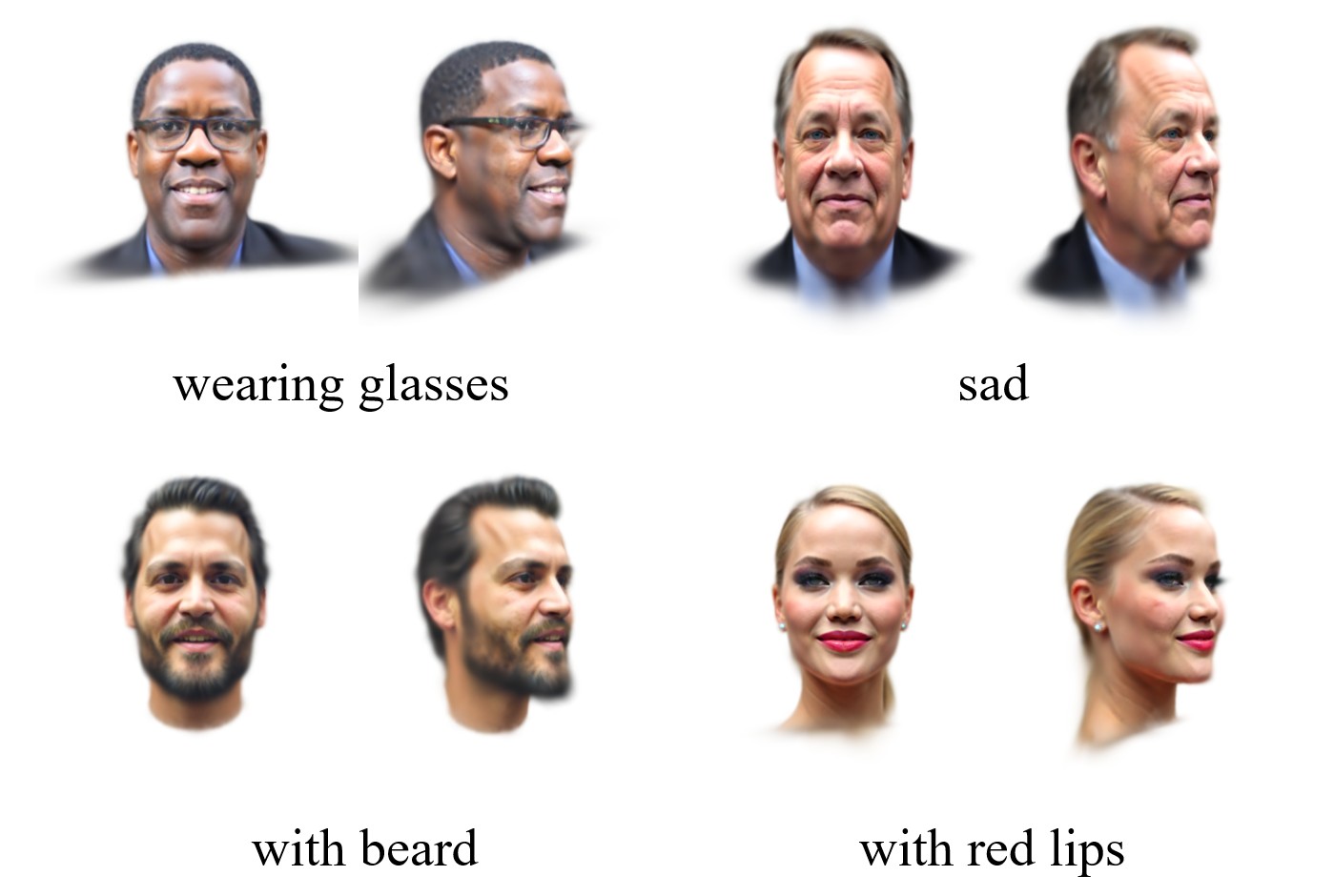}}
\caption{\textbf{Application showcase of SIE3D's expressive generation 
capabilities.} Our method enables fine-grained control over various avatar attributes, 
such as expressions, glasses, and beards, through text prompts.}
\label{fig:res3}
\end{figure}

Our method demonstrates strong expressive generation capabilities. As shown in 
Figure 3, we can generate 3D avatars with various features using different text prompts. 
These results prove the significant potential of our method for downstream 
applications like gaming and virtual reality.

\section{Conclusion}
\label{sec:conclusion}

In this paper, we propose SIE3D, a novel method for generating expressive 
3D avatars from a single image and text description. By introducing a decoupled 
text conditioning mechanism and an expression-aware loss based on a facial 
recognition model, our method achieves fine-grained control over 
expression and appearance attributes while maintaining high-fidelity identity. 
Experimental results show that SIE3D achieves competitive 
performance in identity preservation, semantic control, and expression fidelity.

\vfill\pagebreak

\bibliographystyle{IEEEtran}
\bibliography{strings,refs}

\end{document}